# Deep Galaxy: Classification of Galaxies based on Deep Convolutional Neural Networks

Nour Eldeen M. Khalifa[1,*], Mohamed Hamed N. Taha[1,*], Aboul Ella Hassanien[1,*], I. M. Selim[2,3]

[1]*Faculty of Computers and Information, Cairo University*
[2] *National Research Institute of Astronomy and Geophysics, Cairo, Egypt*
[3]*Computer science Department Integrated Thebes Institutes Cairo, Egypt*
*Scientific Research Group in Egypt, (SRGE)* http://www.egyptscience.net

*Abstract*—**In this paper, a deep convolutional neural network architecture for galaxies classification is presented. The galaxy can be classified based on its features into main three categories Elliptical, Spiral, and Irregular. The proposed deep galaxies architecture consists of 8 layers, and the one main convolutional layer for features extraction with 96 filters, followed by two principles fully connected layers for classification. It is trained over 1356 images and achieved 97.272% in testing accuracy. A comparative result is made, and the testing accuracy was compared with other related works. The proposed architecture outperformed other related works regarding testing accuracy.**

*Index Terms*—galaxies classification, Deep Convolutional Neural Networks, Computational astrophysics.

## I. Introduction

Studying the types and the properties of galaxies are important as it offers important clues about the origin and the development of the universe. The classification of the galaxy is an important role in studying the formation of galaxies and evaluation of our universe. Galaxy morphological classification is a system used to divide galaxies into groups based on their visual appearance. There are several schemes in use by which galaxies can be classified according to their morphologies. Galaxy classification is used to help astrophysicists in facing this challenge. It is done on huge databases of information to help astrophysicists in testing theories and finding new conclusions for explaining the physics of processes governing galaxies, star-formation, and the evaluation of universe [1].

Historically, galaxies classification is a matter of visually inspecting two-dimensional images of galaxies and categorizing them as they appear. Even though expert human classification is somewhat reliable, it is simply too time-consuming for huge amounts of astronomical database taken recently because the increase in the size of telescopes and the CCD camera have has produced extremely large datasets of images, for example, the Sloan Digital Sky Survey (SDSS). These data are too much to analyze manually feasible. Galaxy classification is based on images and spectra. This classification was considered a long-term goal for astrophysicists. However, the complicated nature of galaxies and quality of images have made the classification of galaxies challenging and not accurate [2].

Galaxy classification system helps astronomers in the process of grouping galaxies per their visual shape. The most famous being the Hubble sequence Hubble sequence is considered one of the most used schemes in galaxy morphological classification. The Hubble sequence was created by Edwin Hubble in 1926 [3]. In the past few years, advancements in computational tools and algorithms have started to allow automatic analysis of galaxy morphology. There is several machine learning methods are used to improve the classification of galaxy images.

Prior researchers do not achieve satisfying results. In [8], the authors perform automated morphological galaxy classification based on machine learning and image analysis. They depend on feed-forward neural network and locally weighted regression method for classification. The achieved accuracy was about 91%. In 2013, [9] used naive base classifier and Random Forest classifiers for the morphological galaxy classification. The achieved accuracy was about 91% for Random Forest classifiers and 79% for the Nave base classifier. The authors in [10] proposed a method using the supervised machine learning system based on Non-Negative matrix factorization for images of galaxies in the Zsolt frei Catalog. The achieved accuracy was about 93%. In 2017, [11] proposed a new automated machine supervised learning astronomical classification scheme based on the Nonnegative Matrix Factorization algorithm. The accuracy of the this scheme was about 92%.

Convolution operation is well-known in the computer vision and signals processing community. The convolutional operation is frequently used by conventional computer vision, especially for noise reduction and edge detection [4]. The idea of a Convolutional Neural Network (CNN) is not recent. In 1998, CNN achieved great results for handwritten digit recognition [5]. However, they dramatically drop down due to memory and hardware constraints, besides the absence of large training data. They were unable to scale to much larger images. With the huge increase in the processing power, memory size and the availability of powerful GPUs and large datasets, it was possible to train deeper, larger and more complex models [6]. The machine learning Researchers had been working on learning models which included learning and extracting features from images.

Deep Learning has achieved significant results and a huge improvement in visual detection and recognition with a lot of categories [7]. Raw data images are used be deep learning as input without the need of expert knowledge for optimization of segmentation parameter or feature design.

The rest of the paper is organized as follows. In Section (II) the galaxy dataset characteristic is explained. Section (III) discuss the proposed Deep Galaxies CNN architecture. The experimental results discussed in Section (IV). Finally, Section (V) summarizes the main findings of this paper.

## II. GALAXIES DATA SET CHARACTERISTICS

The dataset used in this research are taken from of the EFIGI catalog [12]. This catalog dataset consists of more than 13,000 images of galaxies and contains samples from different Hubble types of galaxies. The catalog merges data from standard surveys and catalogs (Principal Galaxy Catalogue, Sloan Digital Sky Survey, Value-Added Galaxy Catalogue, HyperLeda, and the NASA Extragalactic Database) [11]. In this research, Hubble galaxies types including Elliptical, Spiral, and Irregular were selected according to the availability of the captured images. The size of the images varies in width and height. Table (I) illustrates the types of galaxies along with the number of training, validation and testing images for each galaxy type while Figure (1) shows three samples from the three type of Galaxy images.

TABLE I
GALAXIES TYPES WITH NUMBER OF TRAINING, VALIDATION AND TESTING

| Galaxy Type | Total Images | Training Set | Validation Set | Testing Set |
|---|---|---|---|---|
| Elliptical | 617 | 370 | 117 | 130 |
| Spiral | 513 | 308 | 97 | 108 |
| Irregular | 216 | 130 | 41 | 45 |

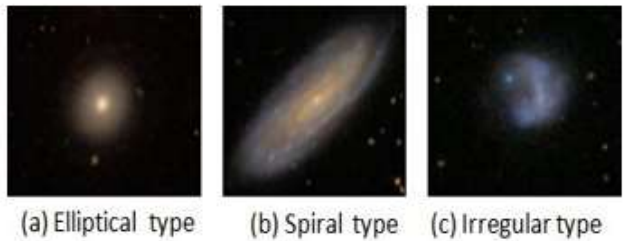

Fig. 1. Sample Galaxy images

### III. THE PROPOSED DEEP GALAXIES ARCHITECTURE

The proposed architecture of the deep network for the galaxies classification is introduced in detail in Figure 2 and Figure 3. Figure 2 illustrates the layers of the proposed architecture, while Figure 2 visualizes the proposed architecture in a graphical representation. It consists of 8 layers, made up of one main convolutional layer for features extraction, followed by two principle fully connected layers for classification. The first layer is the input layer. The second layer is convolution layer. The third layer is a Rectified Linear Unit (ReLU) is which used as the nonlinear activation function. An intermediate pooling is performed in the fourth layer, with subsampling. Then, a fully connected layer has 24 neurons in layer number five, respectively, with ReLU activation function, while the last fully connected layer has three neurons and uses a soft-max layer to obtain class memberships as illustrated in Figure 3.

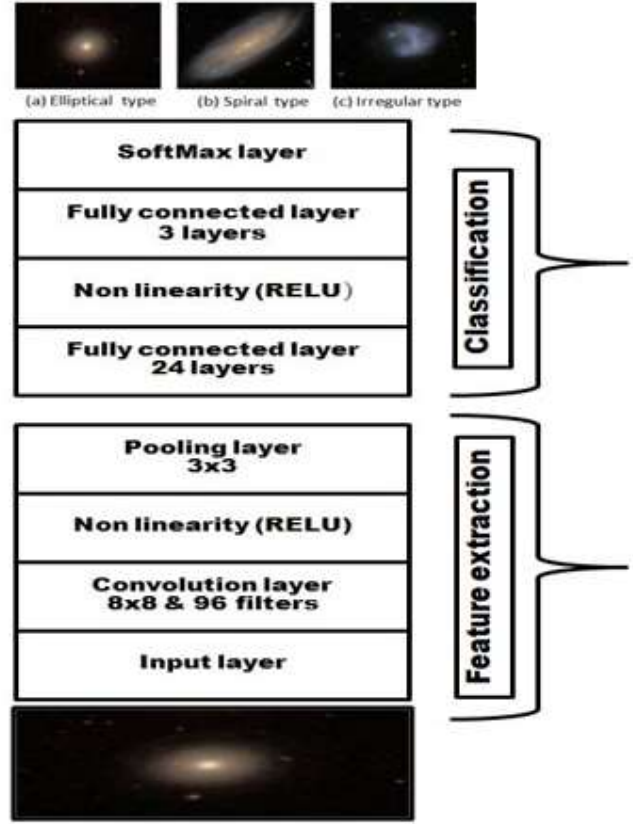

Fig. 2. Detailed layers description for the proposed deep CNN architecture

Visualizing the feature extraction and classification layers in the proposed deep neural architecture will give a better understating. Figure 4 shows the different images resulted from applying first convolution layer with

96 filters and RELU to the input image. In the classification layers, the first fully connected layer and its RELU will produce galaxy images from the passed output images from the first convolution and RELU as illustrated in Figure 5.

## IV. EXPERIMENTAL RESULTS

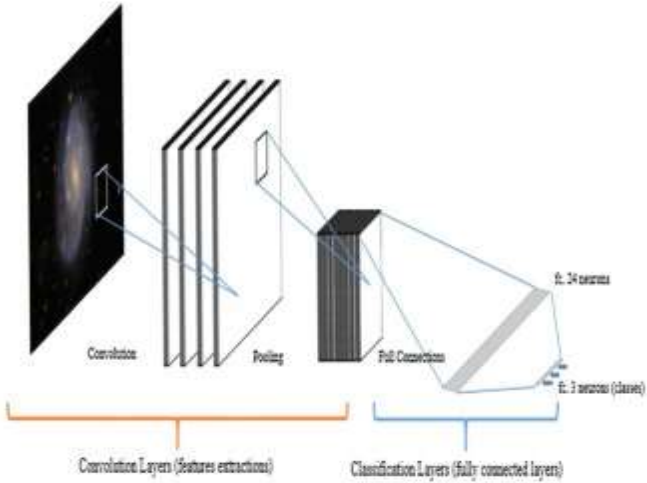

Fig. 3. Abstract view of the proposed deep CNN for galaxies classification

The proposed architecture was implemented using a software package (MATLAB). The implementation was CPU specific. All experiments were conducted on a server with Intel Xeon E5-2620 processor (2 GHz) and 96 GB Ram. To measure the accuracy of the proposed architecture for classifying galaxies types based on deep convolutional neural networks, five different runs were done, and the median accuracy was calculated. Table II represented a comparative result between the proposed architecture and other related works. The proposed architecture-trained over 1346 images. Figure 6 illustrates the training accuracy over some iterations.

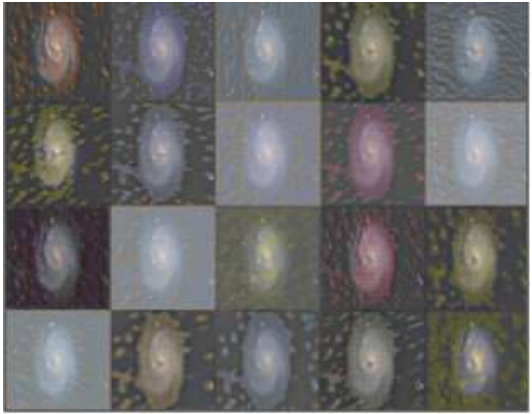

Fig. 4. Typical first convolutional and RELU layer features visualization

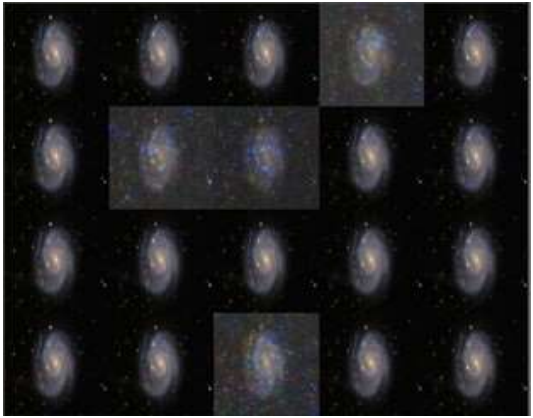

Fig. 5. Typical first fully connected layer visualization.

TABLE II
COMPARATIVE RESULTS FOR GALAXIES CLASSIFICATION

| Related work | Year | Paper Title | Short Description | Accuracy |
|---|---|---|---|---|
| 8 | 2004 | Machine learning and image analysis for morphological galaxy classification | Used neural network and a locally weighted regression for classification | 91% |
| 9 | 2007 | A Hierarchical Model for Morphological Galaxy Classification | Used Naive Bayes classifier, the rule-induction algorithm C4.5 and random forest. | 91.64% |
| 10 | 2016 | Galaxy Image Classification using Non-Negative Matrix Factorization | Used a method based on Non-Negative matrix factorization for images of galaxies in the Zsolt frei Catalog | 93% |
| 11 | 2017 | Automated morphological classification of galaxies based on projection gradient nonnegative matrix factorization algorithm | Used automated machine supervised learning astronomical classification scheme based on the Nonnegative Matrix Factorization algorithm. | 92% |
| Proposed Architecture | 2017 | Deep Galaxy: Classification of Galaxies based on Deep Convolutional Neural Networks | Used Deep Convolutional Neural Networks. | 97.272% |

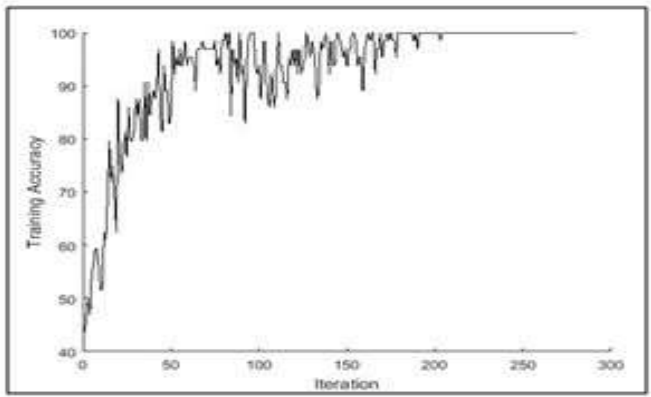

Fig. 6. Training accuracy progress against a number of iteration in the training phase.

## V. CONCLUSIONS AND FUTURE WORKS

The classification of galaxies is considered one of the interesting topics that took the concerns of researchers in last few years. The rise of deep convolutional neural networks science also attracted more researchers into this field. In this paper, a deep convolutional neural network architecture for galaxies classification was introduced. It was trained over 1356 images and achieved 97.272